%% file: root.tex
\documentclass[letterpaper, 10 pt, conference]{ieeeconf}  

\IEEEoverridecommandlockouts                              

\usepackage{graphics} 
\usepackage{epsfig} 
\usepackage{mathptmx}
\usepackage{times}
\usepackage{amsmath}
\usepackage{amssymb}
\usepackage{amsfonts}
\usepackage{pdfpages}
\usepackage{booktabs}
\usepackage{multirow}
\usepackage{subfigure}
\usepackage{threeparttable}
\usepackage{caption}

\renewcommand{\Vec}[1]{\mbf{#1}}
\newcommand{\real}{\mathbb{R}}

\usepackage[capitalise]{cleveref}

\usepackage{color}

\title{\LARGE \bf
Visual-Inertial Multi-Instance Dynamic SLAM with Object-level Relocalisation
}

\author{Yifei Ren$^{1,*}$, Binbin Xu$^{1,*}$,  Christopher L. Choi$^{1}$, and Stefan Leutenegger$^{1,2}$
\thanks{* contributed equally}
\thanks{1 The authors are with Department of Computing, Imperial College London, United Kingdom.
 	{\tt\small \{yifei.ren20, b.xu17, christopher.choi, s.leutenegger\}@imperial.ac.uk}}
\thanks{2 The author is also with the Smart Robotics Lab, Technical University of Munich, Germany}
\thanks{The supplementary video can be watched on: https://youtu.be/6GY5cBwvuJE.}
 	}

\input{notation-math-defs}
\usepackage{amssymb}
\usepackage{amsmath}

\begin{document}

\maketitle
\thispagestyle{empty}
\pagestyle{empty}

\begin{abstract}

In this paper, we present a tightly-coupled visual-inertial object-level multi-instance dynamic SLAM system. Even in extremely dynamic scenes, it can robustly optimise for the camera pose, velocity, IMU biases and build a dense 3D reconstruction object-level map of the environment. Our system can robustly track and reconstruct the geometries of arbitrary objects, their semantics and motion by incrementally fusing associated colour, depth, semantic, and foreground object probabilities into each object model thanks to its robust sensor and object tracking. In addition, when an object is lost or moved outside the camera field of view, our system can reliably recover its pose upon re-observation. We demonstrate the robustness and accuracy of our method by quantitatively and qualitatively testing it in real-world data sequences.
\end{abstract}

\section{INTRODUCTION}
Simultaneous Localisation and Mapping (SLAM) seeks to simultaneously estimate the sensor pose as well as the surrounding scene geometry. Visual SLAM has attracted much attention due to its precision, speed and low-cost. However, most existing SLAM systems have a strong assumption of a static world~\cite{Davison:etal:PAMI2007, Newcombe:etal:ISMAR2011}, where the global environment does not change throughout time. A small number of dynamic objects are eliminated as outliers and are purposefully excluded from the tracking and mapping stages. This setting, however, is not necessarily valid in real world applications, particularly in highly dynamic scenes, where humans are constantly moving and interacting with environments.

To improve the robustness of SLAM systems in a dynamic environment, many authors have proposed various solutions. One direction is to detect and mask out all \emph{potentially} moving elements in a scene, i.e.\ robustly tracking the camera pose against a static background model~\cite{Scona:etal:ICRA2018}. All moving objects are intentionally disregarded and only the background is reconstructed. The detection of moving parts is typically based on inconsistencies between the map and the measurement, and more recently, relies on learning-based semantic segmentation~\cite{Bescos:etal:RAL2018, Barnes:etal:ICRA2018}. Building object-centric maps for each detected object in the scene is another option~\cite{Runz::Agapito::ICRA2017, Runz:etal:ISMAR2018, Xu:etal:ICRA2019, Bescos:etal:RAL2021}. Object-level tracking and mapping can be conducted for each object and the camera pose can be robustly measured against static objects. In this work we approach the dynamic SLAM problem from an object-level, we believe such a system can assist robots in developing object-awareness of their surroundings in the same way that humans do, allowing for a wide range of robotic applications~\cite{Batra:etal:ARXIV2020}. Furthermore, they do not have to rely on \emph{a priori} assumptions on what objects are static, but rather adapt by inferring which specific objects are static at any point in time, therefore they are able to cope with a scene that is largely static, but composed of many potentially moving objects (such as parked cars).

\begin{figure}[t]
    \centering
    \subfigure[Dataset: Book and chair]{
    \begin{minipage}[b]{0.46\linewidth}
        \centering
        \includegraphics[width=0.48\textwidth]{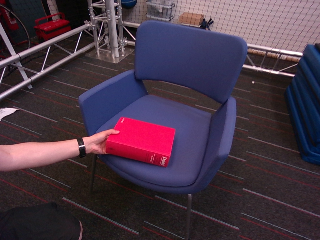}
        \includegraphics[width=0.48\textwidth]{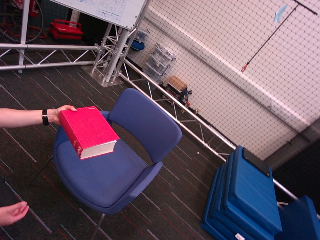}\\
        \includegraphics[width=0.48\textwidth]{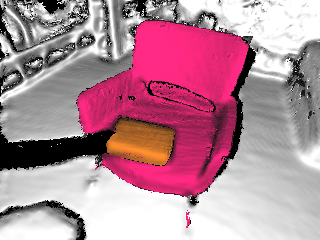}
        \includegraphics[width=0.48\textwidth]{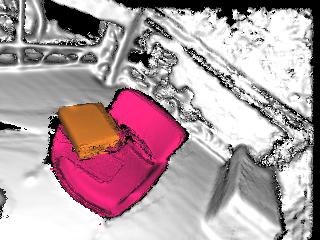}\\
        \includegraphics[width=0.48\textwidth]{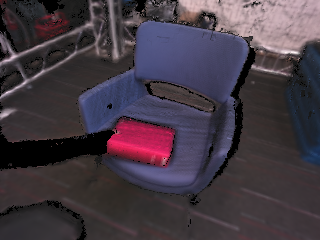}
        \includegraphics[width=0.48\textwidth]{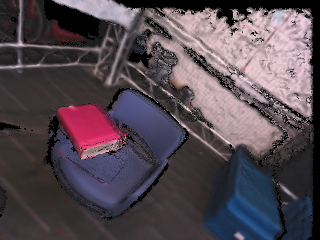}\\
    \end{minipage}
    }
    \subfigure[Dataset: Moving chair]{
    \begin{minipage}[b]{0.46\linewidth}
        \centering
        \includegraphics[width=0.48\textwidth]{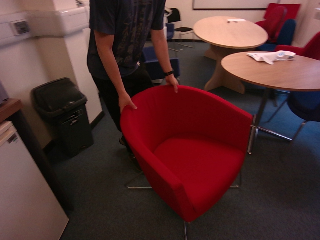}
        \includegraphics[width=0.48\textwidth]{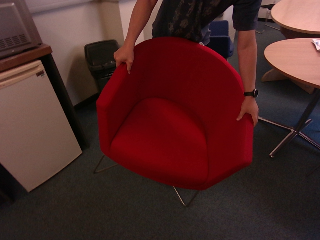}\\
        \includegraphics[width=0.48\textwidth]{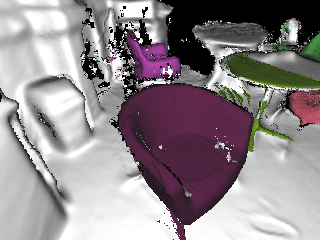}
        \includegraphics[width=0.48\textwidth]{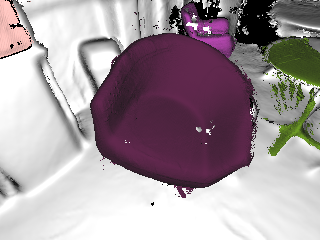}\\
        \includegraphics[width=0.48\textwidth]{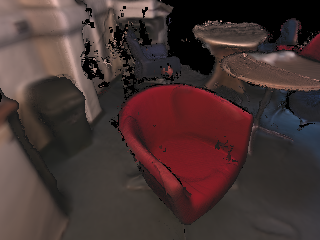}
        \includegraphics[width=0.48\textwidth]{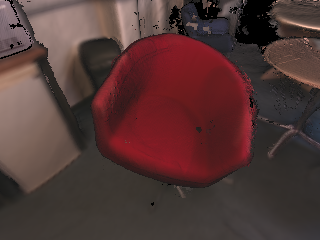}
    \end{minipage}
    }
    \caption{The reconstruction result of our system in dynamic scenes: input RGB image (top row), semantic fusion (middle row) and geometry reconstruction (bottom row).}
    \label{fig: object rotate}
\end{figure}

Despite significant improvement in robustness over static SLAM systems~\cite{Mur-Artal:etal:TRO2017, Vespa:etal:RAL2018}, most object-level SLAM systems assess whether an object is static or dynamic by pre-determining some movable semantic classes~\cite{Bescos:etal:RAL2021} or assume dynamic objects do not occupy the majority of the camera view.~\cite{Xu:etal:ICRA2019}. This assumption is inadequate in many real-world applications such as AR / VR, or self-driving cars. Furthermore, most object-level dynamic SLAM systems are only evaluated in non-repetitive dynamic settings. When an item disappears and reappears, these systems cannot build data association and recover their poses, resulting in duplicate object models. 

In this paper, we aim to improve the robustness of object-level dynamic SLAM systems. To robustly estimate camera pose in a highly dynamic environment, we integrate readings from an Inertial Measurement Unit (IMU) and an RGB-D sensor in a tightly-coupled manner by jointly estimating the camera pose, velocity, and IMU biases. It delivers reliable camera pose estimation even in a highly dynamic scenario while keeping the gravity direction observable, which is important for creating meaningful object-level maps, as shown in \cref{fig: object rotate}. Furthermore, to robustly track object poses in a larger scale, we introduce an object relocalisation step by recovering the previously reconstructed model using keyframe-based feature matching. It builds reliable data association between object-level maps and new observations and reduces object map replications when robots explore a larger-scale environment.

In summary, our main contribution consists of three parts:
\begin{itemize}
\item To be best of our knowledge, this is the first tightly-coupled dense RGB-D-Inertial object-level multi-instance dynamic SLAM system;
\item We propose an object re-localisation method to recover moving objects that disappear and reappear in the camera view;
\item We have extensively tested our system from small-scale to room-scale dynamic scenes, demonstrating the effectiveness and robustness of our proposed method.
\end{itemize}

\section{RELATED WORK}
\paragraph{Object-level dynamic SLAM}
Object-level dynamic SLAM has been regarded as an important topic in the SLAM community since~\cite{Wang:etal:ICRA2003}, as it is critical for robots to understand the surrounding environment which may contain dynamic objects. The goal of such a system is to detect and track dynamic objects while robustly estimating the sensor pose in a dynamic environment. With rapid progress being made in dense visual SLAM~\cite{Newcombe:etal:ISMAR2011, Whelan:etal:IJRR2016}, different approaches to tracking objects has emerged in recent years. The work of~\cite{Runz::Agapito::ICRA2017} for instance, tracks moving objects in the environment using motion residuals, or \cite{Runz:etal:ISMAR2018} where they used instance segmentation to continuously fuse colour and geometric information into an individual map for each object. The object map representation of the above works are based on surfels built on ElasticFusion~\cite{Whelan:etal:IJRR2016}, which cannot provide free space information for downstream robotic applications. Instead, MID-Fusion~\cite{Xu:etal:ICRA2019} leverages a memory efficient octree-based volumetric representation of a Signed Distance Field (SDF) and further conducts semantic fusion for each detected object. Major progress has been made towards improving the robustness and accuracy of these systems. EM-Fusion~\cite{Strecke:Stuckler:CVPR2019}, for example, proposes to estimate object pose by directly align the object SDF with the input frame. DetectFusion~\cite{Hachiuma:etal:BMVC2019}, on the other hand, combines 2D object detection and 3D geometric segmentation to detect and segment the motion of semantically unknown objects. 

Similar to visual static SLAM, instead of alternating the optimisation of tracking and mapping as most dense SLAM systems do, as explained in~\cite{Engel:etal:PAMI2017}, another direction is to formulate a joint probabilistic inference on map and pose for higher object tracking accuracy~\cite{Durrant-Whyte:etal:RAM2006}, with the caveat of sacrificing the dense map representation. DynaSLAM-II~\cite{Bescos:etal:RAL2021} represents objects as sparse pointclouds and jointly optimises the camera pose, object poses and geometries in a pose graph optimisation.  ClusterSLAM~\cite{Huang:etal:CVPR2019} formulates object detection and tracking as a clustering of landmark movement and solves it as a batch optimization problem. Following that, they reformulated it as an online VO SLAM that also considers semantic detection~\cite{Huang:etal:CVPR2020}. 

Previous work has explored using a motion model to track objects, such as~\cite{Runz:etal:ISMAR2018, Xu:etal:ICRA2019}, for example, where objects are tracked using a zero velocity motion model. The work of ~\cite{Bescos:etal:RAL2021}, on the other hand, uses a constant velocity motion model, or ~\cite{Barfoot:book2017,Huang:etal:CVPR2020} which uses a white-noise-on-acceleration prior. These motion assumptions, however, may be inadequate in describing the true motion of any arbitrary object, especially when the objects disappear and reappear in the camera view. In this work, we forgo the need for a motion prior, and formulate the object relocalisation pipeline with a combination of a keyframe-based feature matching and dense residual verification stage.

\paragraph{Visual-inertial odometry and SLAM}
Estimating robot pose by fusing both IMU and camera data has gain popularity in recent years, as it has proven to be robust and accurate in many robotic applications. There are generally two approaches to VIO/VI-SLAM, loosely-coupled and tightly-coupled systems. Early VIO systems estimated robot poses from IMU and camera measurements independently and fused together at a later stage, this loosely-coupled approach has the benefit of reducing computing complexity e.g.~\cite{Konolige:etal:RR2011}. With improved processing capability, however, recent state-of-the-art VIO systems have adopted a tightly-coupled approach that jointly optimises all states by taking into account all correlations among them, leveraging both the IMU and camera measurements simultaneously, e.g.~\cite{Mourikis:Roumeliotis:ICRA2007, Leutenegger:etal:IJRR2014}.

Another strategy to classify visual-inertial estimation problems is to assess if they are solved using recursive filtering or batch non-linear optimisation methods. Recursive filtering methods update states from visual data and only use IMU measurements for state propagation. MSCKF~\cite{Mourikis:Roumeliotis:ICRA2007} is the first work introducing a tightly-coupled filtering-based visual-inertial odometry. ROVIO~\cite{Bloesch:etal:IROS2015} similarly adopts a tightly-coupled filtering approach, but with direct photometric residuals instead of indirect feature associations. Nonlinear optimisation-based approaches jointly optimise visual error term and together with integrated IMU measurements. By restricting the optimisation to a limited sliding window of keyframes via marginalisation, OKVIS~\cite{Leutenegger:etal:IJRR2014} achieves high accuracy while maintaining real-time speed. OKVIS2~\cite{Leutenegger:ARXIV2022} further extends it by constructing pose graph factors from marginalised observations and achieving loop closure. ORB-SLAM III~\cite{Campos:etal:TRO2021} is also a recent real-time VI-SLAM system that, however, drops past estimation uncertainties when fixing old states.

These sparse VIO/VI-SLAM systems show robust and accurate estimation in various environments. However, the reconstructed maps are too sparse for safe robot
navigation and meaningful scene understanding. There have been several efforts at providing visual-inertial dense mapping. Extending from ElasticFusion~\cite{Whelan:etal:IJRR2016}, VI-ElasticFusion~\cite{Laidlow::etal::IROS2017} presents a dense RGB-D-Inertial SLAM system with map deformations. Kimera~\cite{Rosinol:etal:ICRA2020} also creates a dense mesh reconstructions with a VIO-frontend and provides a pose-graph optimisation backend used upon loop closure. It can also provide semantically annotated map for scene understanding.

Even if dense VIO/VI-SLAM systems can densely reconstruct a global scene, they typically lack awareness of the objects in the scene.
A number of studies have looked into visual-inertial object-level SLAM.
Using inertial measurements, scale and gravity direction become observable. VI object-level SLAM can provide both scale discrimination and global orientation for visual recognition~\cite{Dong:etal:CVPR2017} and semantic mapping~\cite{Fei:Soatto:ECCV2018}. The sensor pose estimation can also be further improved by including the object pose and shape into a tightly-coupled filtering VIO system~\cite{Shan:etal:IROS2020}.

However, most existing VIO/VI-SLAM systems above only target a static environment, despite the fact that IMU provides reliable measurements of sensor ego-motion. Moving objects are carefully excluded from the estimated states. A few recent works extend them into dynamic scenes, but the objects are mainly composed of sparse 3D landmarks \cite{Qiu:etal:TRO2019, Eckenhoff:etal:ICRA2020}. In contrast, this work fuses IMU with RGB-D sensor measurements in a tightly-coupled nonlinear optimisation manner to provide robust sensor tracking, even in extreme dynamic scenarios, while dense tracking and reconstructing each detected object in the scene. We further provide an object relocalisation step to manage the re-visited moving objects.

\section{SYSTEM OVERVIEW}
\begin{figure*}[tb]
    \centering
    \includegraphics[width=0.88\linewidth]{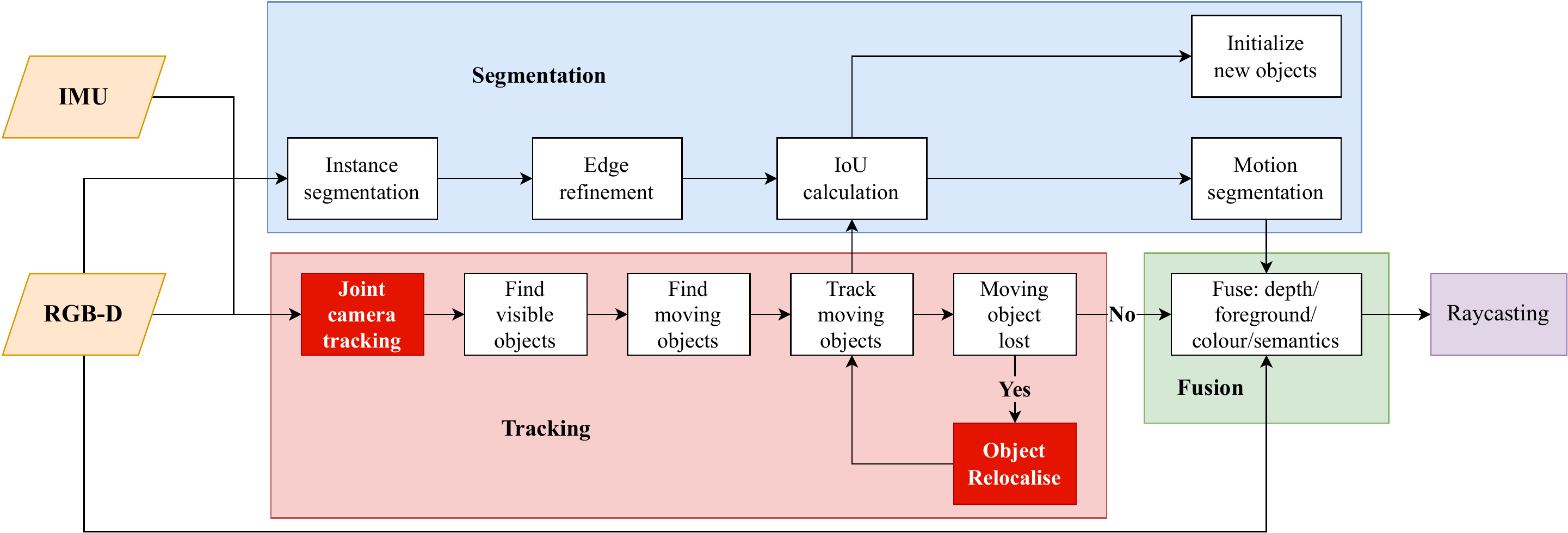}
    \caption{Pipeline of VI-MID system.}
    \vspace{-1em}
    \label{fig: pipeline}
\end{figure*}
 
\begin{figure}[!t]
    \centering
    \includegraphics[width=\linewidth]{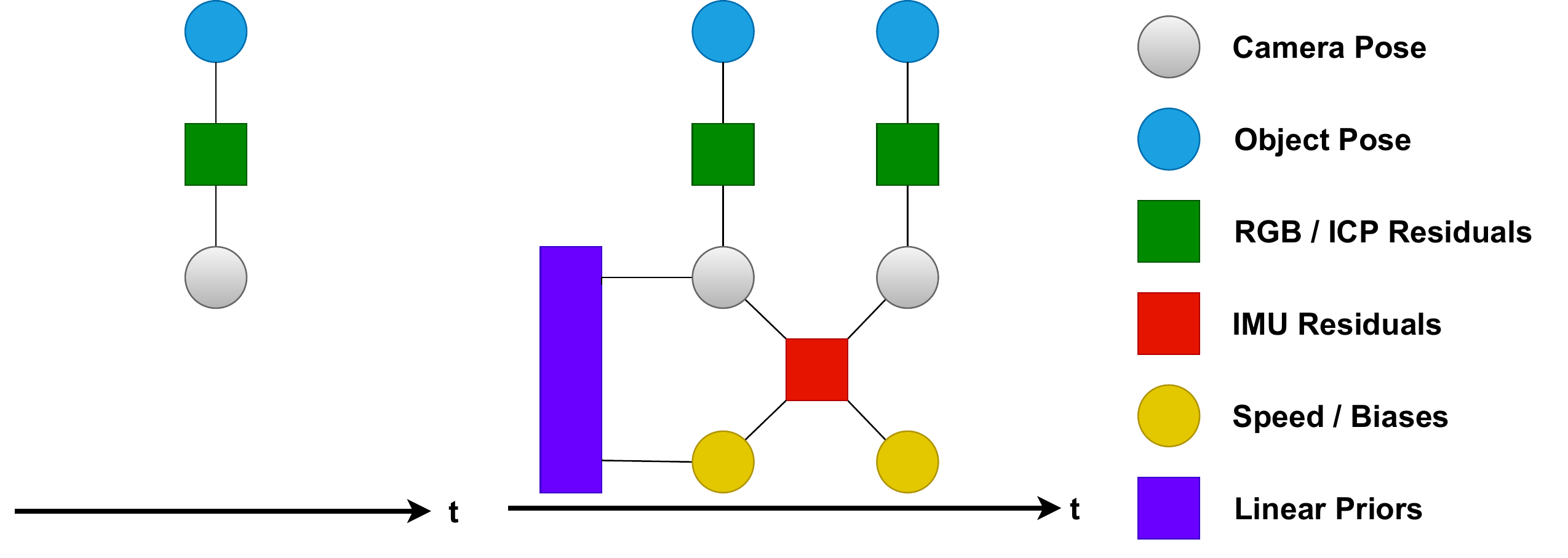}
    \caption{The factor graphs of the state variables and measurements involved in the MID-Fusion (left) versus VI-MID (right): incorporating inertial measurements introduces temporal constraints, and augments the state by speed and IMU biases.}
    \vspace{-1em}
    \label{fig: factor graph}
\end{figure}

Our VI-MID system builds upon the octree-based object-level multi-instance dynamic SLAM system MID-Fusion \cite{Xu:etal:ICRA2019}. \cref{fig: pipeline} shows the pipeline of our refined system, which consists of four major parts: tracking, segmentation, fusion and raycasting. In the tracking part, the camera pose is estimated by jointly optimizing the photometric, geometric and IMU residuals. During the system's operation dynamic objects in the environment are detected and tracked, with their poses refined using photometric and geometric residuals. Lastly, the fusion and raycasting parts are identical to MID-Fusion \cite{Xu:etal:ICRA2019}.

\section{NOTATION}

We employ the following notation throughout this work: The position vector and rotation matrix from the origin of the world frame, $\cframe{W}$, to the origin of the camera frame, $\cframe{C}$, be denoted as $\pos{W}{C}$ and $\Crot{W}{C}$ respectively; the corresponding Hamiltonian quaternion is written as $\q{W}{C}$. A homogeneous transformation matrix representing the camera pose in the world frame is written as $\T{W}{C}$, where it can also be used to transform homogeneous points observed from the camera frame $\cframe{C}$ to the world frame $\cframe{W}$. Furthermore, current and previous RGB-D data are distinguished as live (L) and reference (R) respectively. For example, the current RGB-D measurement contains the intensity image frame $I_L$ and depth map $D_L$, with $\Vec{u}_{L}$ denoting a single 2D pixel location in the live measurement with $[\cdot]$ used to perform pixel lookup (including bilinear interpolation).

Three different coordinate frames will be used in this work:
\begin{itemize}
    \item{$\cframe{W}$, the world frame, corresponding to the the first camera frame position with orientation aligned with gravity.}
    \item{$\cframe{C}$, the camera frame in which RGB-D data is observed.}
    \item{$\cframe{S}$, the sensor frame in which IMU data is observed.}
\end{itemize}
The extrinsic offset between camera and sensor frame is calibrated beforehand.
Temporal offsets between the camera and IMU are assumed to be synchronized and negligible.

\section{TRACKING MODULE}

\subsection{States}

Upon arrival of a new RGB-D measurement, the tracking module estimates the position of the RGB camera expressed in world frame $\pos{W}{C} \in \real^{3}$, the RGB camera orientation $\q{W}{C} \in S^{3}$, the velocity of the IMU in sensor frame $_{S}\Vec{v}_{WS} \in \real^{3}$, and finally the gyroscope and accelerometer biases $\Vec{b}_{g} \in \real^{3}$ and $\Vec{b}_{a} \in \real^{3}$ respectively. The system state vector $\Vec{x}$ for a specific time instance is therefore:

\begin{align}
    \Vec{x} 
        := &\left[
            {_{W}\Vec{r}_{C}^T},
            \enspace \Vec{q}_{WC}^T, 
            \enspace {_{S}\Vec{v}_{WS}^{T}},
            \enspace \Vec{b}_g^T,
            \enspace \Vec{b}_a^T
        \right]^T \\
    \nonumber
    &\in \real^{3}
    \times S^{3} 
    \times \real^{9}. 
\end{align}

In general the system state exists on a manifold and we defined a group operator $\boxplus$ to perturb the states in tangent space, such that $\Vec{x} = \bar{\Vec{x}} \boxplus \delta\Vec{x}$ around the estimate $\bar{\Vec{x}}$ with a local perturbation $\delta\mathbf{x}$. Vector space quantities such as $_{W}\Vec{r}_{C}$, ${_{S}\Vec{v}_{WS}}$, $\Vec{b}_g$ and $\Vec{b}_a$ were updated via standard vector addition. In contrast, $\Vec{q}_{WC}$ was updated by a combination of the group operator (quaternion multiplication) and exponential map (including the the map into the Lie algebra) was used such that $\Vec{q} \boxplus \delta{\boldsymbol{\alpha}} = \mathrm{Exp}(\delta\boldsymbol{\alpha}) \otimes \Vec{q}$. As a result, the minimal local coordinate representation is:
\begin{equation}
    \delta\mathbf{x} 
        = \left[
            \delta\mathbf{r}^T, 
            \enspace \delta\boldsymbol\alpha^T, 
            \enspace \delta\mathbf{v}^T, 
            \enspace \delta\mathbf{b}_g^T, 
            \enspace \delta\mathbf{b}_a^T 
        \right]^T 
        \in \enspace \real^{15}.
\end{equation}
Similarly, we also defined a group operator $\boxminus$ to invert the above $\boxplus$ operation. Further details and a more detailed treatment of differential calculus can be found at~\cite{Leutenegger:etal:IJRR2014} and~\cite{Bloesch:etal:CoRR2016}.

\subsection{RGB-D-Inertial Sensor Tracking}

The tracking problem aims to estimate the reference state $\Vec{x}_R$ and live state $\Vec{x}_L$. This is solved by minimising the cost function $E_{\text{track}}$ which consists of four terms. The photometric error (RGB) $E_{\text{photo}}$, dense point-to-plane ICP error (Depth) $E_{\text{ICP}}$, inertial error (IMU) $E_{\text{inertial}}$, and marginalisation priors $E_{\text{prior}}$ in a non-linear optimisation framework. 
\cref{fig: factor graph} visualises the factor graph of this tracking cost function and compares the one used in MID-Fusion~\cite{Xu:etal:ICRA2019}. Each error term contains the sum of residuals $e$, weighted by their respective measurement uncertainty $w$, except for the marginalisation priors where we are using the same marginalisation scheme as~\cite{Laidlow::etal::IROS2017}, and both $\mathbf{H}^{*}$ and $\mathbf{b}^{*}$ are priors obtained by marginalising out the reference state $\Vec{x}_R$,
\begin{align}
  &E_{\text{track}}(\Vec{x}_R, \Vec{x}_L) = \;
      E_{\text{photo}}
      + E_{\text{ICP}}
      + E_{\text{inertial}} 
      + E_{\text{priors}}, \\ 
  \nonumber \\
  &E_{\text{photo}} = 
        \sum_{u_R \in M_R}
        w_p \; \rho(e_p), \\
  &E_{\text{ICP}} = 
        \sum_{u_L \in M_L}
        w_g \; \rho(e_g), \\
  &E_{\text{inertial}} = 
        \; w_s \; \rho(\|\mbf{e}_s\|), \\
  &E_{\text{priors}} = 
    \dfrac{1}{2}
    \left(
        \Vec{x}_R \boxminus \bar{\Vec{x}}_R - 
        \mathbf{H}^{* -1}
        \mathbf{b}^{*}
    \right)^{T}
    \mathbf{H}^{*}
    \left(
        \Vec{x}_R \boxminus \bar{\Vec{x}}_R - 
        \mathbf{H}^{* -1}
        \mathbf{b}^{*}
    \right),
\end{align}
where $\rho(\cdot)$ is a robust kernel (Cauchy loss function was used), and $M_R, M_L$ are masks for excluding invalid measurements for both RGB and depth data due to invalid correspondences, occlusions and dynamic objects. 

\subsubsection{Dense Photometric Residual}
The photometric residual aims to minimise the intensity difference of the reference pixel $\Vec{u}_R$ observed in both the reference intensity frame $I_R$ and live intensity frame $I_L$, and is defined as follows:
\begin{align}
    e_{p}&(\T{W}{C_{L}}) =
        I_R \left[ \Vec{u}_{R} \right]
        -
        I_L
        \left[
            \pi \left(
                \T{W}{C_L}^{-1} 
                \T{W}{C_R} \; \pi^{-1}(\Vec{u}_{R}, \; D_{R}[\Vec{u}_R])
            \right)
        \right],
\end{align}
where $\pi(\cdot)$ is the projection function that maps a 3D point observed in the camera frame onto the image plane, $\pi(\cdot)^{-1}$ performs the opposite and back-projects an image point back to a 3D point. $\T{W}{C_L}$, $\T{W}{C_R}$ are the live and reference RGB camera poses respectively, and $D_{R}$ is the \textit{rendered} reference depth map. This is in direct contrast to the standard photometric residual that uses the latest depth measurement available. This decision increases tracking robustness when raw input depth is not avaiable, e.g. when the camera is too close to a wall or a surface.

\subsubsection{ICP Residual}
In addition to the photometric residual, using the approach proposed in \cite{Newcombe:etal:ISMAR2011} the ICP residual adds additional geometric constraints between RGB-D measurements by minimizing the point-plane depth residual between the live depth map and the \textit{rendered} reference depth map:
\begin{equation}
    e_{g}(\T{W}{C_{L}}) = 
    {{}_W\textbf{n}^{r}[\Vec{u}_{R}]} 
    \cdot \left(
    \T{W}{C_L} \;
        {}_{C_L} \Vec{v} [\Vec{u}_L] \;
        - \; 
        {}_{W}\Vec{v}^{r}[\Vec{u}_{R}]
    \right),
\end{equation}
where $_{C_L}\textbf{v}$ is the live vertex map in the camera coordinate by back-projection, $_W\textbf{v}^r$ and $_W\textbf{n}^r$ are the rendered vertex map and normal map in the world coordinate. For each pixel $_L\textbf{u}$ in the live frame, its corresponding pixel $\Vec{u}_R$ in the reference frame can be calculated using projection and back-projection:
\begin{equation}
    \Vec{u}_{R} = \pi \left(
        \T{W}{C_R}^{-1} \;
        \T{W}{C_{L}} \;
            \pi^{-1}(\Vec{u}_R, \enspace D_R[\Vec{u}_R])
        \right).
\end{equation}

\subsubsection{Measurement Uncertainty Weight}

In order to weigh the contribution of the RGB and Depth sensor towards the estimated RGB camera pose $\T{W}{C}$ fairly, the measurement uncertainty of each sensor is required. For the RGB camera, the measurement uncertainty is assumed to be uniform and constant across all pixels. For the depth sensor, the depth measurement quality is related to the structure of the RGB-D sensor and the depth range. Using the inverse covariance definition for depth measurement uncertainty in~\cite{Laidlow::etal::IROS2017}, given the baseline $b$, disparity $d$, focal length $f$, and the uncertainties in the x-y plane $\sigma_{xy}$ and disparity direction $\sigma_{z}$, the standard deviation $\sigma_{D}$ for depth measurements in $x, \; y, \; z$ coordinates can be modelled as:
\begin{equation}
    \sigma_D = \left(
        \frac{{D}_L[\Vec{u}_L]}{f} \sigma_{xy}\;, \enspace
        \frac{{D}_L[\Vec{u}_L]}{f} \sigma_{xy}\;, \enspace
        \frac{{D}^2_L[\Vec{u}_L]}{fb} \sigma_z
    \right).
\end{equation}
As a result, the weight for ICP residuals using the inverse covariance of measurement uncertainty is:
\begin{equation}
    w_{g}= \frac{1}{(_W\textbf{n}^r)^T {_W\textbf{n}^r} {\sigma_{D}^T} \sigma_{D}^{}}.
\end{equation}

\subsubsection{Inertial Residual}
To make our system robust to inadequate vision or depth data, we adopt the approach of~\cite{Laidlow::etal::IROS2017} and pre-integrate IMU measurements numerically between the reference and live time instance in the inertial residual term to estimate both the reference and live RGB camera poses, as well as the speed and IMU biases. The inertial residual term is given by:
\begin{equation}
    \mbf{e}_{S} = \hat{\Vec{x}}_{L}(\mathbf{x}_{R}) \boxminus \Vec{x}_{L},
\end{equation}
where $\Vec{x}_R$ is the reference state, $
\hat{\Vec{x}}_L$ is the prediction of the live state propagated by IMU measurements from the reference state.


\subsection{Object Tracking}

To handle objects in the environment, we adopted the approach of MID-Fusion~\cite{Xu:etal:ICRA2019} by estimating the relative pose between the i-th object $O_i$ and RGB camera pose $C$ in the live frame $\T{C_{L}}{O_{i}}$. This part is computed separately from the sensor tracking. First, object masks are extracted using Mask R-CNN, followed by geometric and motion refinement. Secondly, the associated photometric and ICP residual terms are parametrised in an object-centric form to estimate $\T{C_{L}}{O_{i}}$. The ICP residuals associated with the object are modified to align the \textit{live} vertex map in the live object frame with the \textit{rendered} vertex map in the reference object frame:
\begin{equation}
    {e}_{g}(\Vec{T}_{C_{L}O_{i}}) = 
        \Crot{W}{O_{i}}^{-1} 
        {{}_{W}\textbf{n}^{r}[\Vec{u}_{R}]} 
        \cdot 
        (
            \T{C_{L}}{O_{i}}^{-1}
            \; {}_{C_{L}}\Vec{v}[\Vec{u}_L] 
            \; - \; \T{W}{O_{i}}^{-1} 
            \; {}_{W}\Vec{v}^{r}[\Vec{u}_R]
        ).
    \label{eq:obj_1}
\end{equation}
Analogously, the photometric residuals are re-parameterised to:
\begin{align}
    {e}_{p}(\T{C_{L}}{O_{i}}) &=  
        I_R\left[ \Vec{u}_{R} \right] 
        - I_L\left[
            \pi \left(
           \Vec{u}_L 
            \right)
        \right] 
        \;,
    \label{eq:obj_2}
\end{align}
where $\Vec{u}_L$ is now
\begin{align}
    \Vec{u}_L &= 
        \T{C_{L}}{O_{i}} 
        \T{C_{R}}{O_{i}}^{-1}
        \pi^{-1}(\Vec{u}_{R}, \; D_R[\Vec{u}_{R}]).
\end{align}
The above object-centric residuals are then optimised in a three-level coarse-to-fine scheme using the Gauss-Newton algorithm.

\subsection{Object Re-localisation}
\begin{figure}[tb]
    \centering
    \includegraphics[width=0.85\linewidth]{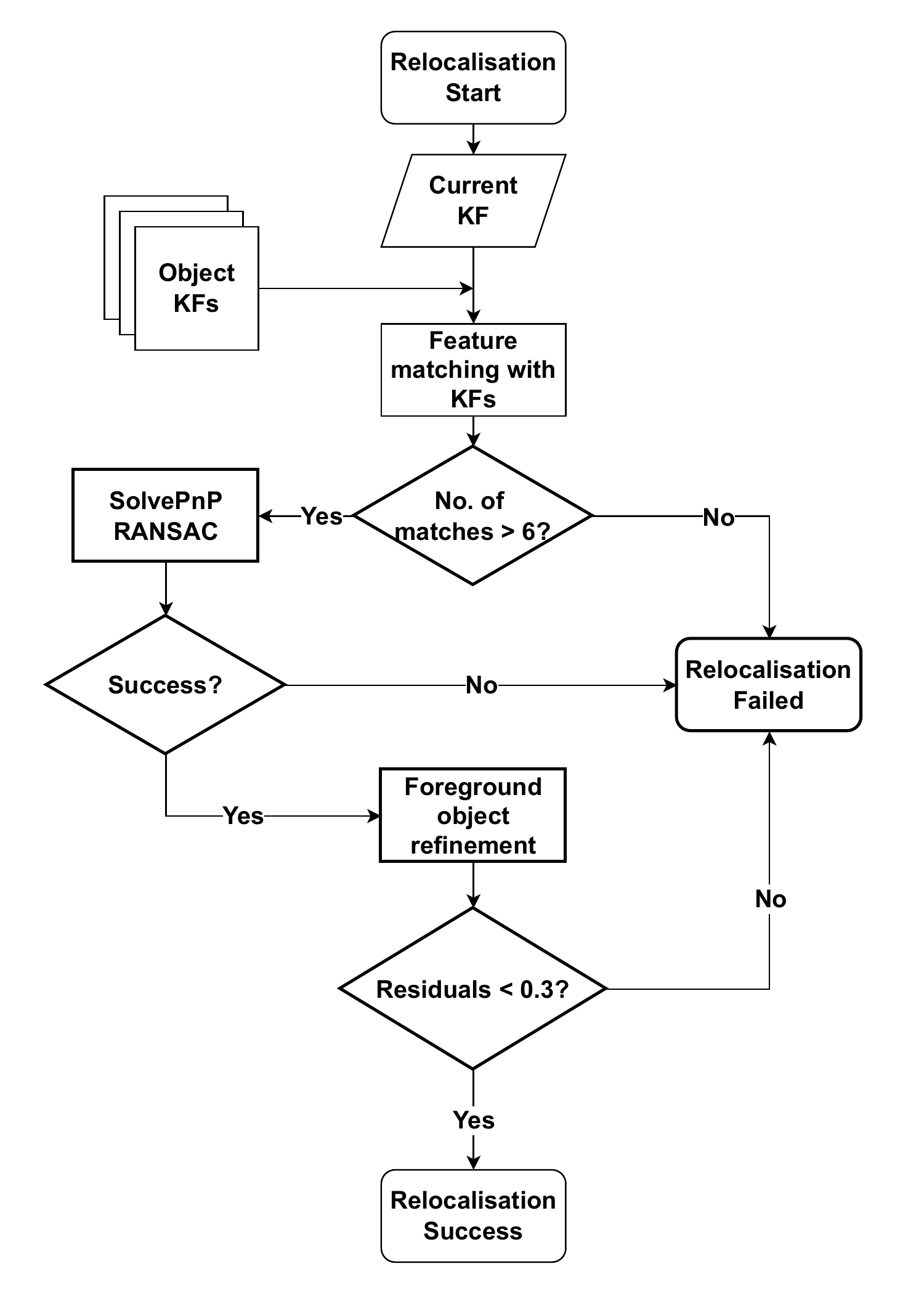}
    \caption{Object Re-localisation Pipeline.}
    \vspace{-1em}
    \label{fig:reloc_pipeline}
\end{figure}

In many practical applications, objects in the environment are not always static nor observable by the RGB-D sensor at all times, making it challenging when previously lost objects are re-observed at a later time. Previously existing work did not address this issue. In this work, we propose a keyframe based object re-localisation method to handle such cases. 

Similar to MID-Fusion~\cite{Xu:etal:ICRA2019}, we perform an object-level raycasting to associate object detection with existing models. For each unmatched object, we initialise a new object model in an octree-based volumetric representation \cite{Vespa:etal:RAL2018}. Its object coordinate w.r.t.\ the world frame $\T{W}{O_{i}}$ is set to the identity rotation and its translation component is obtained through depth measurements from $D_L$. When one object is matched with an existing model with IoU larger than 0.8, we integrate the colour, depth, and semantic probability into its model. Object keyframes associated with the object are also created with detected BRISK~\cite{Leutenegger:etal:ICCV2011} features on it. New keyframes are only added when view-change larger than 10$^{\circ}$ is reached. Each object keyframe contains the relative pose between the camera and the object, $\T{C}{O_{i}}$, image keypoints $\Vec{z} \in \real^{2}$, feature descriptors and object points in the camera coordinate $_{C}\Vec{p} \in \real^{3}$ associated with each keypoint.

In the event that an object goes out of the camera view, it is marked as lost and re-localisation on each subsequent camera frame is required. To save on computational resources, the relocalisation pipeline is only triggered for each unmatched detection whose center of the object is not close to the image boundary. This is implemented in the form of a re-localisation threshold $R(O_{i})$ and is defined as:
\begin{equation}
    R(O_{i}) = \dfrac{S_{\text{current}}(O_{i})}{S_{\text{max}}(O_{i})}
    \label{eq:relocalisation_threshold}
\end{equation}
where $S_{\text{current}}(O_{i})$ is the current object mask size and $S_{\text{max}}(O_{i})$ is max object mask size. Empirically we found $R(O_{i}) > 0.7$ yields the best results.

When a lost object is re-observed and the relocalisation threshold \cref{eq:relocalisation_threshold} is met, the relocalisation pipeline (See \cref{fig:reloc_pipeline}):
\begin{enumerate}
    \item{iterates through previous object keyframes to obtain the object's keypoints and feature descriptors,}
    \item{performs feature matching on the object features against features detected in the current frame,}
    \item{finds the best match within all the object keyframes,}
    \item{perform SolvePnp on the best matched features to obtain the relative pose on the detected object $\T{C_R}{C_L}$, which can be used to obtain $\T{C_L}{O_{i}}$,}
    \item{Object pose $\T{W}{O_{i}}$ is further refined using \cref{eq:obj_1} and \cref{eq:obj_2}. If the residual is too large, the object matching is deemed invalid,}
    \item{Once the object pose is recovered, we delete the new duplicate object model and retrieve the previously reconstructed object model}.
\end{enumerate}

\section{EXPERIMENTAL RESULTS}
\subsection{Experimental Setup}
To validate our proposed system we collected two types of datasets. The first type was collected using Vicon motion capture system to provide ground-truth: this set was used to evaluate the tracking performance of our system. The second set was collected in an office environment with dynamic objects: this set was used to evaluate the handling and relocalisation of dynamic objects in the environment. All datasets were recorded using an Intel RealSense D455 which has an RGB-D (15Hz) and IMU (200Hz) sensor. The measurements of all sensors were time-stamped and synchronised.

We compared VI-MID against three other state-of-the-art systems, MID-Fusion \cite{Xu:etal:ICRA2019}, VI-ElasticFusion (VI-EL-F) \cite{Laidlow::etal::IROS2017} (our own implementation of their tracking part), and OKVIS \cite{Leutenegger:etal:IJRR2014} on real world datasets. All the hyperparameters, including ours, are pretuned and fixed throughout all experiments. 
All experiments were conducted on a Linux system with an Intel Core i7-10875H CPU at 2.30GHz with 32GB memory. 

\subsection{Tracking Performance Evaluation}
To evaluate the tracking performance, we compared our system on 5 Vicon datasets (VD) using the absolute trajectory (ATE) root-mean-square error (RMSE) metric. Among the 5 VDs, VD-1 and VD-2 are static environments; VD-3, VD-4 and VD-5 are dynamic with different motions of chair (by human operation), which is the major object in the scene. Meanwhile, all of the five datasets contain 1300-1700 frames. It is worth mentioning that OKVIS is the only sparse feature-based system and does not provide object-level maps.

\begin{figure}
    \centering
    \includegraphics[width=\linewidth]{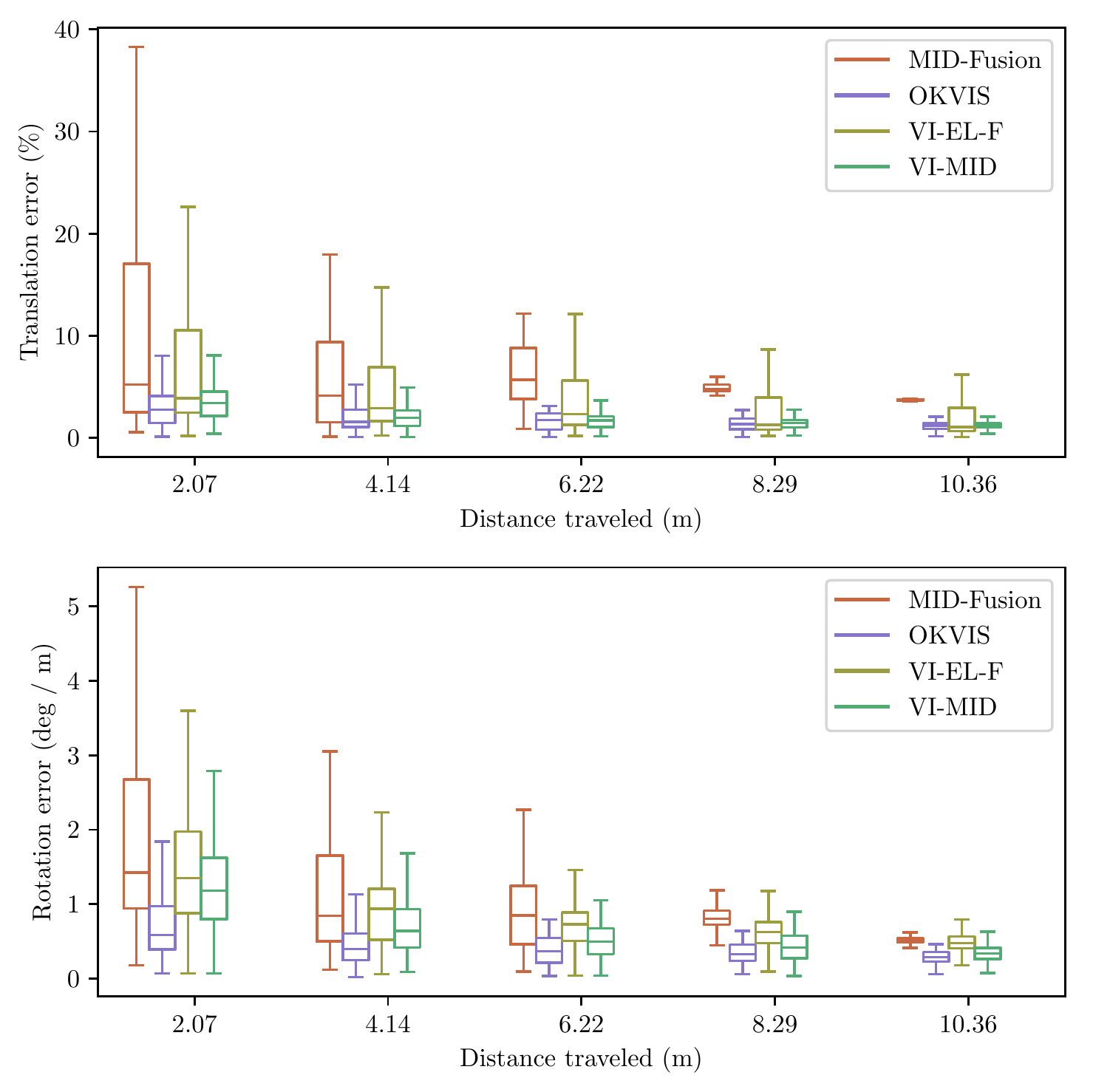}
    \caption{Drift analysis of MID-Fusion, OKVIS, VI-EL-F and VI-MID on the VD-3 seqeunce. VI-MID has the lowest drift over time, while MID-Fusion is not able to finish the trajectory.}
    \vspace{-1em}
    \label{fig:drift}
\end{figure}

In a static environment all systems performed comparably, with VI-MID on average having a lower ATE RMSE trajectory error. However, in a dynamic environment, VI-MID beats both MID-Fusion and VI-EL-F due to VI-MID's ability to handle dynamic objects in the environment (see \cref{tab: ate table} and \cref{fig:drift}). MID-Fusion was unable to complete the tracking on any of the dynamic sequences as it tends to drift when the object that takes up the majority of the view moves at the same time as the camera. In contrast, our tracking method provides reliable tracking performance even in fast dynamic settings. Our system currently does not achieve the same level of tracking accuracy as OKVIS \cite{Leutenegger:etal:IJRR2014}, which jointly optimises the trajectory and (static) sparse map~-- it does not estimate any dense geometry, objects nor object motion.

\begin{table}[tbp]
\footnotesize
    \centering
    \caption{Comparison of ATE RMSE between five different Vicon datasets.}
    \begin{threeparttable}
        \begin{tabular}{ccccc||c}
             \toprule
             Type & Dataset & VI-EL-F & MID-Fusion & \textbf{VI-MID} & OKVIS\\
             \midrule
             \multirow{2}{*}{Static}  & VD-1 & 0.081 & 0.087   & \textbf{0.068} & 0.064 \\
                                      & VD-2 & \textbf{0.055} & 0.083  & 0.058 & 0.024 \\
             \midrule
             \multirow{3}{*}{Dynamic} & VD-3 & 0.174 & (0.342) & \textbf{0.060} & 0.022 \\
                                      & VD-4 & 0.105 & (0.488) & \textbf{0.094} & 0.027 \\
                                      & VD-5 & 0.496 & (0.374) & \textbf{0.239} & 0.081 \\
             \bottomrule
        \end{tabular}
        \begin{tablenotes}
            \footnotesize               
            \item (*) means the tracking cannot be completed.  
        \end{tablenotes}
    \end{threeparttable}
    \label{tab: ate table}
    \vspace{-1em}
\end{table}

\subsection{Object Reconstruction}
\begin{figure*}[!t]
    \centering
    \subfigure[Input sequence]{\includegraphics[width=0.11\linewidth]{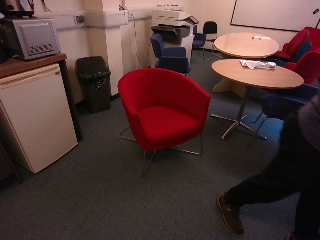}
    \includegraphics[width=0.11\linewidth]{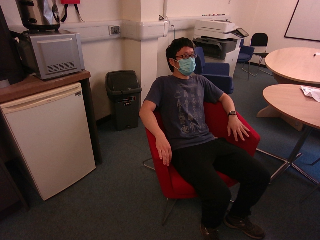}
    \includegraphics[width=0.11\linewidth]{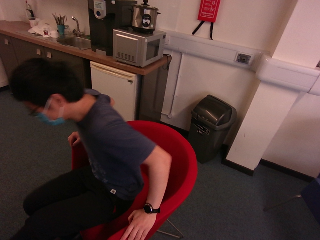}
    \includegraphics[width=0.11\linewidth]{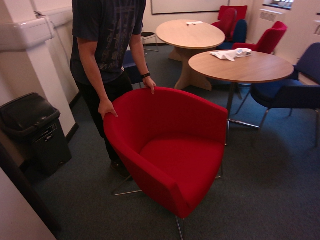}
    \includegraphics[width=0.11\linewidth]{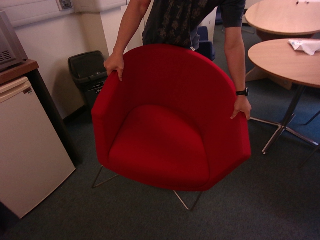}
    \includegraphics[width=0.11\linewidth]{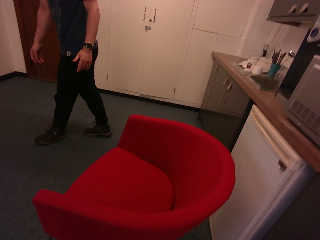}
    \includegraphics[width=0.11\linewidth]{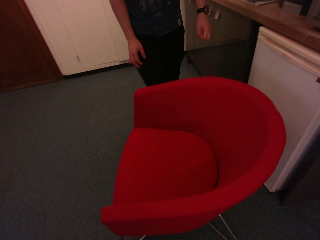}
    \includegraphics[width=0.11\linewidth]{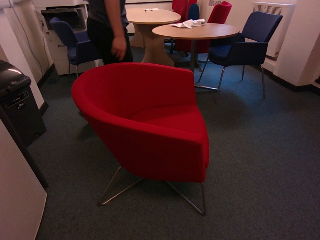}
    }
    \subfigure[Side view of reconstruction]{\includegraphics[width=0.45\linewidth]{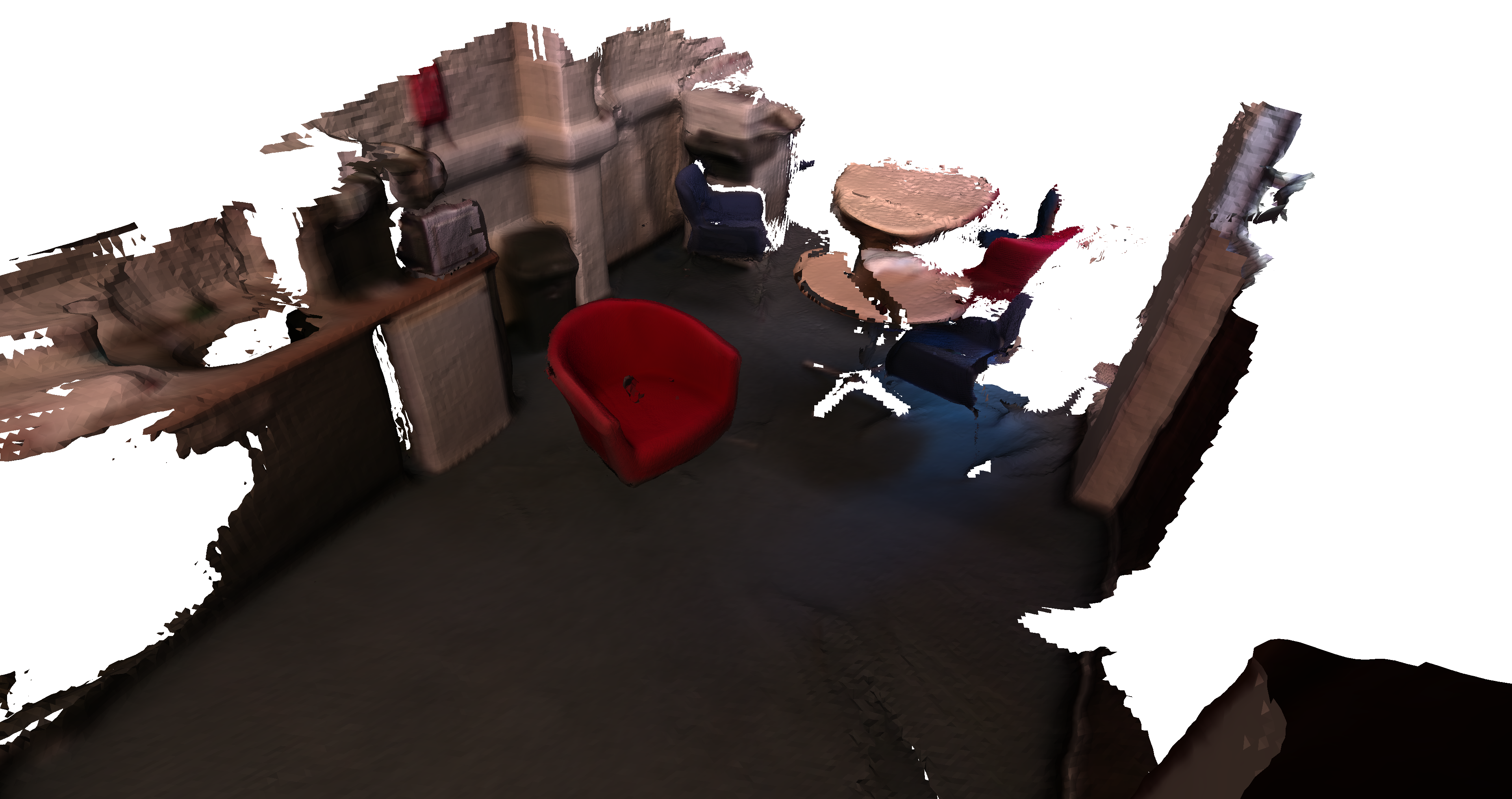}}
    \subfigure[Top view of reconstruction]{\includegraphics[width=0.45\linewidth]{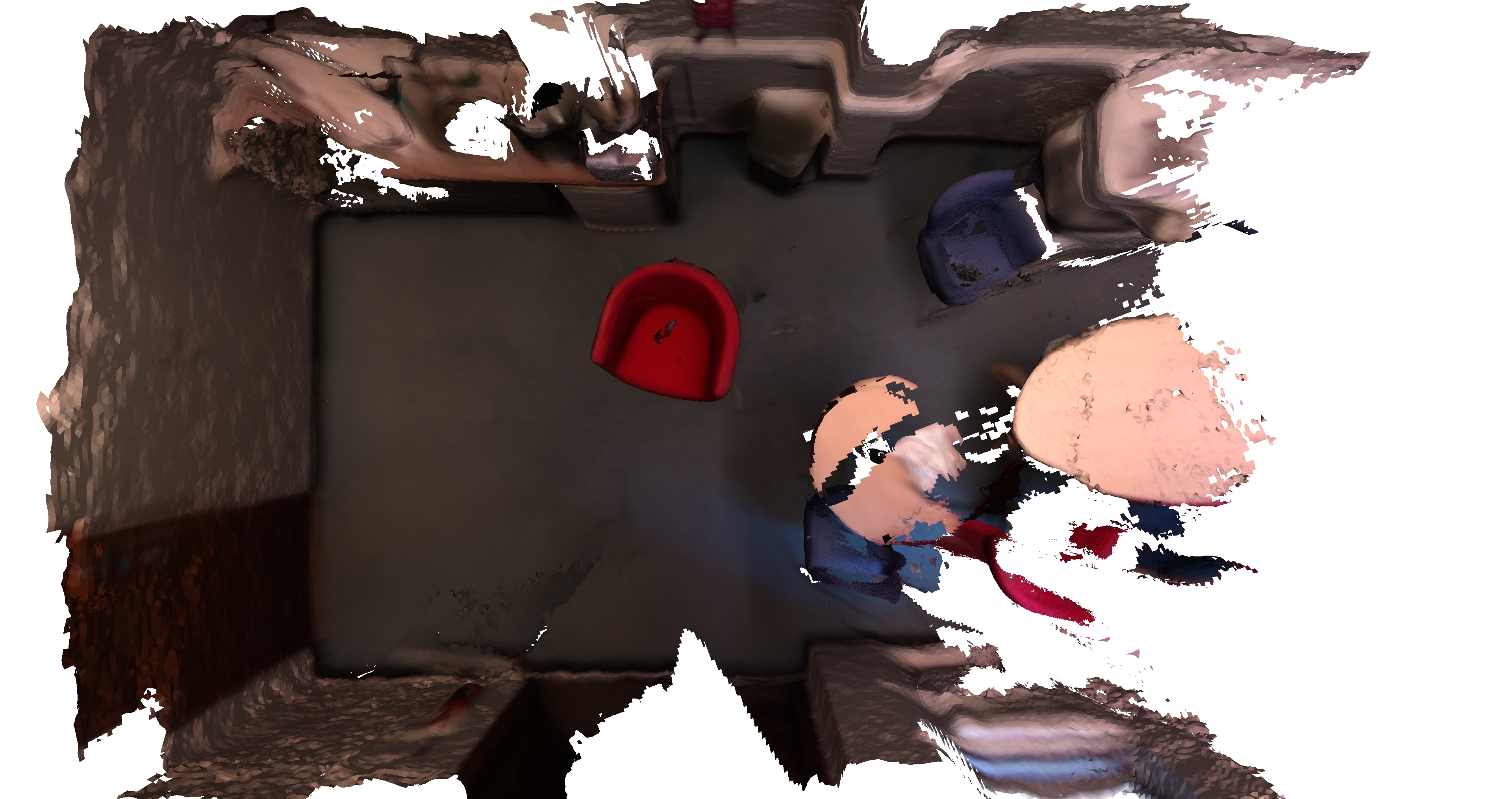}}
    \vspace{-0.5em}
    \caption{Qualitative demonstration of room-scale reconstruction.}
    \label{fig: kitchen recon}
\end{figure*}

To highlight VI-MID's ability to handle dynamic objects, we collected office sequences (762 frames) with dynamic objects to qualitatively demonstrate our system's ability to reconstruct tracked objects and compare our results with MID-Fusion. \cref{fig: kitchen recon} shows the reconstruction of the room. From the sequence of \cref{fig: kitchen recon} (a), the camera constantly changes its angle of view with a person sitting, leaving and moving the chair. However, our VI-MID system could still reconstruct the whole scene reliably, with the correct final position of the chair.

\begin{figure}[tbp]
    \centering
    \subfigure{\includegraphics[width=0.3\linewidth]{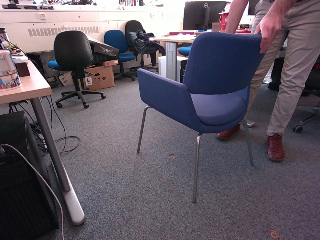}}
    \subfigure{\includegraphics[width=0.3\linewidth]{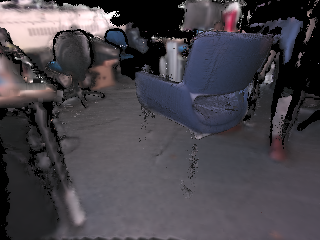}}
    \subfigure{\includegraphics[width=0.3\linewidth]{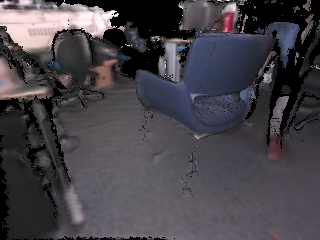}}
    \setcounter{subfigure}{0}
    \subfigure[Input]{\includegraphics[width=0.3\linewidth]{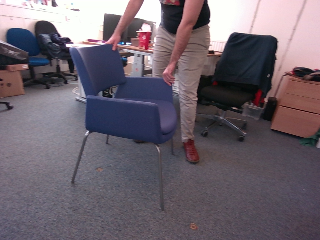}}
    \subfigure[MID-Fusion]{\includegraphics[width=0.3\linewidth]{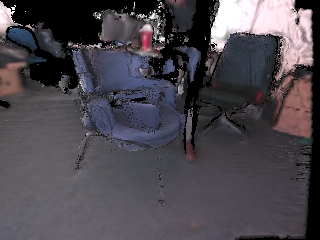}}
    \subfigure[VI-MID]{\includegraphics[width=0.3\linewidth]{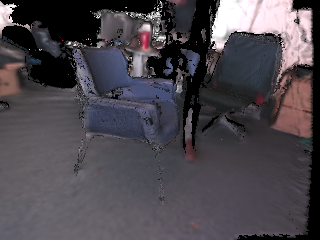}}

    \caption{Qualitative comparison between MID-Fusion and VI-MID: input RGB image (left column), MID-Fusion reconstruction (middle column) and VI-MID reconstruction (right column).}
    \vspace{-1em}
    \label{fig: object recon chair}
\end{figure}

\cref{fig: object recon chair} shows the difference between VI-MID and MID-Fusion in reconstructing a chair before and after it was moved by a person. 

\subsection{Object Relocalisation} 

\begin{figure}[tbp]
    \centering
    \subfigure[Object position before moving]{
        \label{fig: object reloc a}
        \includegraphics[width=0.45\linewidth]{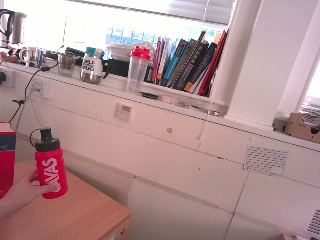}
    }
    \subfigure[Object position after moving]{
        \label{fig: object reloc b}
        \includegraphics[width=0.45\linewidth]{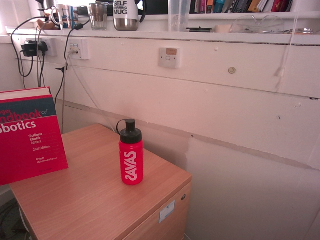}
    }
    \subfigure[Reconstruction without object relocalisation]{
        \label{fig: object reloc c}
        \includegraphics[width=0.45\linewidth]{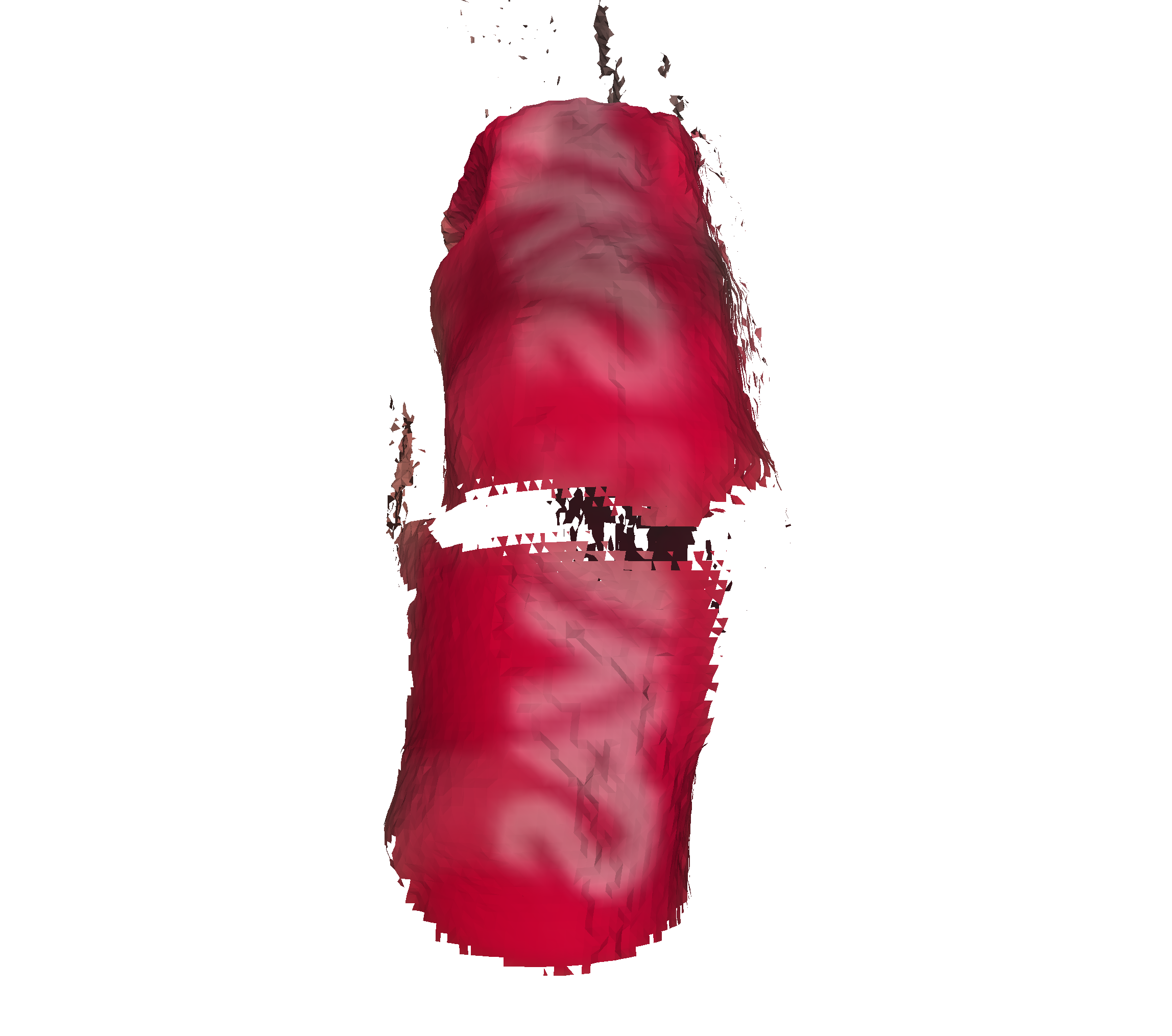}
    }
    \subfigure[Reconstruction with object relocalisation]{
        \label{fig: object reloc d}
        \includegraphics[width=0.45\linewidth]{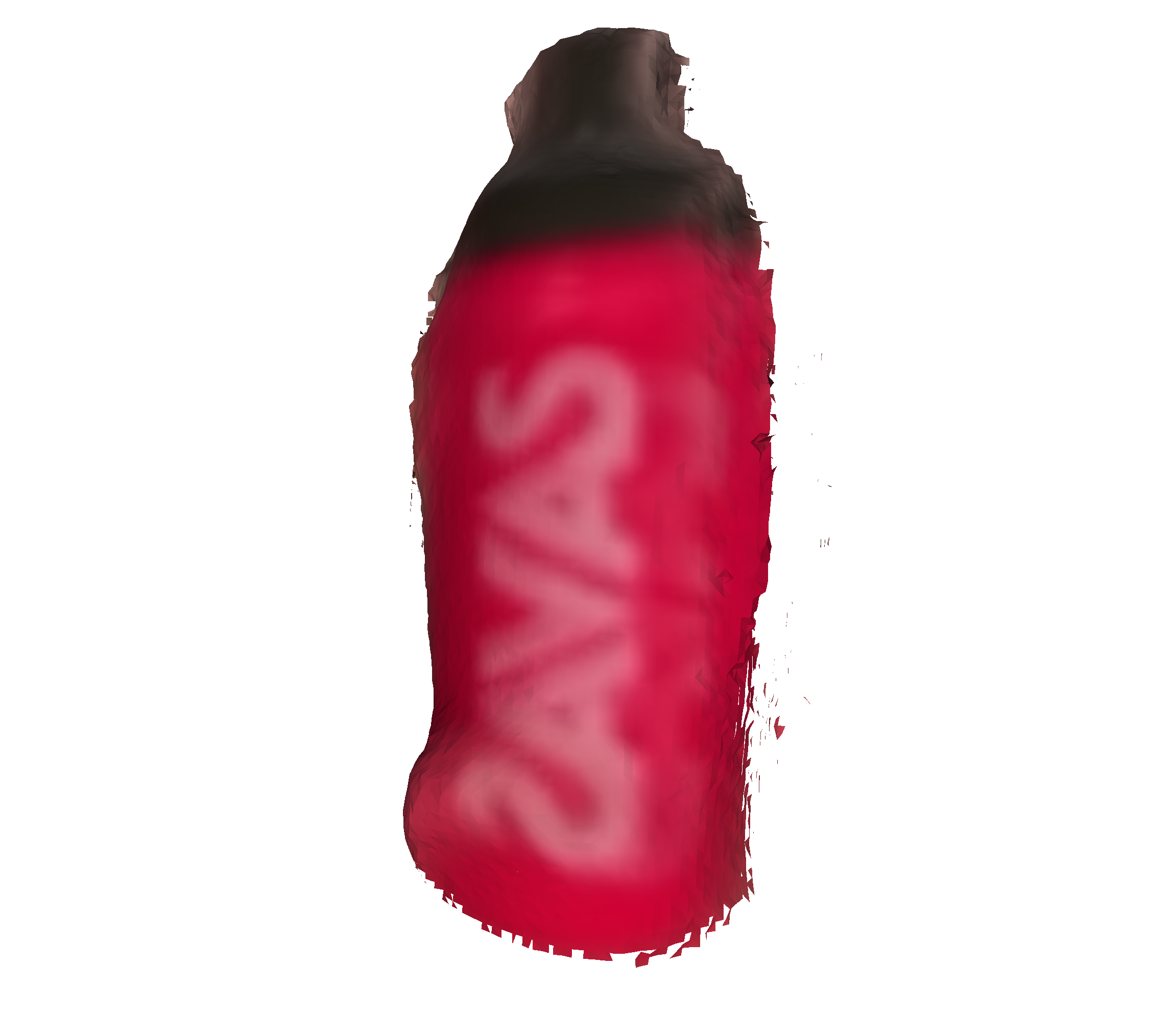}
    }
    \caption{Qualitative comparison between reconstruction of bottle with and without object reconstruction.}
    \label{fig: object reloc}
\end{figure}

The result of object relocalisation is shown in \cref{fig: object reloc}. As seen in \cref{fig: object reloc a} and \cref{fig: object reloc b} the position of the bottle changes while the RGB-D sensor turns away from the bottle and then back. Systems such as MID-Fusion that has no object relocalisation mechanism, would not be able to associate with the same object model, causing a new object model to be created and deleted every time the object goes in / out of camera view, thus making the system inefficient and sub-optimal. On the contrary, with VI-MID's object relocalisation mechanism, \cref{fig: object reloc c} show cases the scenario where the current mask associates with an old rendered mask. And \cref{fig: object reloc d} shows the result of object relocalisation, where the correct pose of the bottle can be estimated and the model of the bottle can continue to be used without any occlusion. 

%

\subsection{Runtime Analysis}
To evaluate the average run-time of each component in VI-MID, we ran our system on 5 data sequences with a varying number of dynamic objects in the environment. The number of moving objects (MO) affects the tracking time of VI-MID. While visible objects (VO) affect segmentation, integration and raycasting times. \cref{tab: times} reports  a detailed break down of average processing time per frame for each component.
\begin{table}[tbp]
\footnotesize
    \centering
    \caption{Run-Time analysis of system components (ms).}
    \begin{tabular}{cccc}
         \toprule
         \textbf{Components} & \textbf{Camera Track} & \textbf{Object Track} & \textbf{Relocalisation}\\
         \midrule
         \textbf{Time (ms)} & 20/VO & 43/MO & 17/KF\\
         \midrule
         \textbf{Components} & \textbf{Segmentation} & \textbf{Integration} & \textbf{Raycasting}\\
         \midrule
         \textbf{Time (ms)} & 9/VO & 8/VO & 5/VO\\
         \bottomrule
    \end{tabular}
    \vspace{-1em}
    \label{tab: times}
\end{table}

\subsection{Discussions}
We have conducted several experiments on real world datasets comparing VI-MID against state-of-the-art dense slam systems such as MID-Fusion and VI-EL-F, where we have shown VI-MID out-performs both MID-Fusion and VI-EL-F in handling dynamic objects in the environment. Additionally, we have qualitatively demonstrated VI-MID's ability to track and relocalise dynamic objects previously tracked in the environment. OKVIS, despite showing better tracking accuracy, it cannot generate an object-level 3D dense representation.  One limitation of VI-MID is that its processing time would scale with the number of objects in the scene, just as in related RGB-D dense object-level SLAM systems~\cite{Runz:etal:ISMAR2018, Xu:etal:ICRA2019}. This limits its real-time performance when the scene becomes complex. 
Besides, if a pre-detected object is lost, we will try to re-localise it indefinitely until it is recovered. This is an implementation we would like to revisit as part of future work.

\section{CONCLUSIONS}
We have introduced a novel system which performs visual-inertial dense SLAM capable of handling multiple dynamic objects in the environment. Our system has three major advantages. First, compared to traditional vision-only dense SLAM, the integration of an IMU adds additional robustness in tracking, specifically in dynamic environments. Second, the system is capable of mitigating the effects of intermittent object detection from Mask R-CNN, since the system can fall back on rendered masks obtained from the object-level map to continually track objects in the environment.

\section*{ACKNOWLEDGMENT}
We thank Tristan Laidlow and Dimos Tzoumanikas for their fruitful discussions. Binbin Xu holds a China Scholarship Council-Imperial Scholarship. This research is supported by Imperial College London, Technical University of Munich, EPSRC grant ORCA Stream B - Towards Resident Robots, and the EPSRC grant Aerial ABM EP/N018494/1.



\bibliographystyle{IEEEtran}
\bibliography{robotvision}

\end{document}

%% file: notation-math-defs.tex
\usepackage{xifthen}
\usepackage{xparse}
\usepackage{subdepth}
\usepackage{leftidx}
\usepackage{bm}
\usepackage{amssymb,amsmath}
\usepackage{amsfonts}

\newcommand{\bbm}{\begin{bmatrix}}
	\newcommand{\ebm}{\end{bmatrix}}

\DeclareMathAlphabet{\mbf}{OT1}{ptm}{b}{n}
\newcommand{\mbs}[1]{{\bm{#1}}}

\newcommand{\mbsbar}[1]{{\overline{\boldsymbol{#1}}}}
\newcommand{\mbshat}[1]{{\hat{\boldsymbol{#1}}}}
\newcommand{\mbstilde}[1]{{\tilde{\boldsymbol{#1}}}}

\newcommand{\mbsdot}[1]{{\dot {\boldsymbol{#1}}}}

\newcommand{\mbfbar}[1]{{\overline{\mbf{#1}}}}
\newcommand{\mbfhat}[1]{{\hat{\mbf{#1}}}}
\newcommand{\mbftilde}[1]{{\tilde{\mbf{#1}}}}

\newcommand{\mbfdot}[1]{{\dot{\mbf{#1}}}}

\newcommand{\cframe}[1]{{\smash{\protect\underrightarrow{\mathcal{F}}_{#1}}}}

\DeclareMathAlphabet{\mathbfit}{OML}{cmm}{b}{it}
\newcommand{\homo}[1]{{\mathbfit{#1}}}

\newcommand{\mbfh}[1]{{\homo{#1}}}

\newcommand{\trans}[3]{\leftidx{_{#1}}{\mbf r}{\IfValueTF{#2}{_{#2#3\hspace{2pt}}}{}}} 

\newcommand{\pos}[2]{\leftidx{_{#1}}{ \mbf r}{_{#2}}} 

\newcommand{\vel}[3]{\leftidx{_{#1}}{\mbf v}{\IfValueTF{#2}{_{#2#3\hspace{2pt}}}{}}} 
\newcommand{\veltilde}[3]{\leftidx{_{#1}}{\mbftilde v}{\IfValueTF{#2}{_{#2#3\hspace{2pt}}}{}}} 
\newcommand{\velbar}[3]{\leftidx{_{#1}}{\mbfbar v}{\IfValueTF{#2}{_{#2#3\hspace{2pt}}}{}}} 
\newcommand{\velhat}[3]{\leftidx{_{#1}}{\mbfhat v}{\IfValueTF{#2}{_{#2#3\hspace{2pt}}}{}}} 
\newcommand{\veldot}[3]{\leftidx{_{#1}}{\mbfdot v}{\IfValueTF{#2}{_{#2#3\hspace{2pt}}}{}}} 

\newcommand{\acc}[3]{\leftidx{_{#1}}{\mbf a}{\IfValueTF{#2}{_{#2#3\hspace{2pt}}}{}}} 
\newcommand{\acctilde}[3]{\leftidx{_{#1}}{\mbftilde a}{\IfValueTF{#2}{_{#2#3\hspace{2pt}}}{}}} 
\newcommand{\accbar}[3]{\leftidx{_{#1}}{\mbfbar a}{\IfValueTF{#2}{_{#2#3\hspace{2pt}}}{}}} 

\newcommand{\rotvel}[3]{\leftidx{_{#1}}{\mbs \omega}{\IfValueTF{#2}{_{#2#3\hspace{2pt}}}{}}} 
\newcommand{\rotveltilde}[3]{\leftidx{_{#1}}{\mbstilde \omega}{\IfValueTF{#2}{_{#2#3\hspace{2pt}}}{}}} 
\newcommand{\rotvelbar}[3]{\leftidx{_{#1}}{\mbsbar \omega}{\IfValueTF{#2}{_{#2#3\hspace{2pt}}}{}}} 
\newcommand{\rotvelhat}[3]{\leftidx{_{#1}}{\mbshat \omega}{\IfValueTF{#2}{_{#2#3\hspace{2pt}}}{}}} 
\newcommand{\rotveldot}[3]{\leftidx{_{#1}}{\mbsdot \omega}{\IfValueTF{#2}{_{#2#3\hspace{2pt}}}{}}} 

\newcommand{\Crot}[2]{\leftidx{}{\mbf C}{_{#1#2\hspace{2pt}}}} 
\newcommand{\T}[2]{\leftidx{}{\mbfh T}{_{#1#2\hspace{2pt}}}} 
\newcommand{\q}[2]{\leftidx{}{\mbf q}{_{#1#2\hspace{2pt}}}} 
